\documentclass[journal]{IEEEtran}

\ifCLASSINFOpdf
\else
   \usepackage[dvips]{graphicx}
\fi
\usepackage{url}

\usepackage{graphicx}
\usepackage{cite}
\usepackage{hyperref}
\usepackage{amsthm,amsmath,amssymb}
\usepackage{mathrsfs}
\usepackage{booktabs}
\usepackage{bm,balance}
\hypersetup{hidelinks}
\usepackage[normalem]{ulem}
\useunder{\uline}{\ul}{}

\usepackage{multirow}
\usepackage{tabularx}
\usepackage{makecell}
\usepackage{colortbl}

\usepackage{amssymb}
\makeatletter

\newcommand{\Rmnum}[1]{\expandafter\@slowromancap\romannumeral #1@}
\makeatother

\usepackage{tikz}
\definecolor{lime}{HTML}{A6CE39}
\DeclareRobustCommand{\orcidicon}{%
    \begin{tikzpicture}
    \draw[lime, fill=lime] (0,0) 
    circle [radius=0.16] 
    node[white] {{\fontfamily{qag}\selectfont \tiny ID}};    \draw[white, fill=white] (-0.0625,0.095) 
    circle [radius=0.007];    \end{tikzpicture}
    \hspace{-2mm}}
\foreach \x in {A, ..., Z}{
    \expandafter\xdef\csname orcid\x\endcsname{\noexpand\href{https://orcid.org/\csname orcidauthor\x\endcsname}{\noexpand\orcidicon}}
}

\begin{document}

\title{Enhance Multi-Scale Spatial-Temporal Coherence for Configurable Video Anomaly Detection}

\author{Kai Cheng, Xinzhe Li, and Lijuan Che

\thanks{(Kai Cheng and Xinzhe Li contributed equally to this work and should be considered co-first authors.) (Corresponding author: Lijuan Che.)}
\thanks{Kai Cheng is with Fudan University, Shanghai, China.}
\thanks{Xinzhe Li is with East China University of Science and Technology, Shanghai, China.}
\thanks{Lijuan Che is with the School of Artificial Intelligence in Traditional Chinese Medicine, the Shanghai University of Traditional Chinese Medicine, Shanghai 201203, China (e-mail: 0000001401@shutcm.edu.cn).}}

\markboth{Journal of \LaTeX\ Class Files, Vol. 14, No. 8, August 2015}
{Shell \MakeLowercase{\textit{et al.}}: Bare Demo of IEEEtran.cls for IEEE Journals}
\maketitle

\begin{abstract}
    The development of unsupervised Video Anomaly Detection (VAD) relies on technologies in the field of signal processing. Since the anomaly is quite ambiguous and unbounded, different detection demands may often be raised even in one scenario. Thus, we propose to design the configurable VAD with flexible solutions targeting to solve the issue that previous methods have to train their models from scratch and waste resources when detection demands even change slightly. Moreover, we also design a dataset with good compatibility to evaluate the VAD performance when changes happen in detection demands. Besides, videos contain important information regarding continuous changes in the object's appearance and motion. Thus, we also propose a module to establish the multi-scale spatial-temporal coherence, which improves the accuracy and has the ability to dynamically adjust and accurately capture spatial-temporal normal patterns. Experiments show that our method not only models coherence effectively but also has better configurable ability.
\end{abstract}


\IEEEpeerreviewmaketitle

\section{Introduction}

\IEEEPARstart{V}{ideo} Anomaly Detection (VAD) is urgently needed for free manual inspection in various fields as a challenging task in the signal processing community. However, acquiring abnormal data is far more difficult. And, the objective of VAD is to detect any anomaly that violates normal patterns rather than in an anomaly set preset manually. Thus, with the development of unsupervised learning, we can utilize frame prediction as the proxy task to detect anomalies by formulating VAD as a one-class classification task. Recently, some deep learning-based methods \cite{luo2021future, lv2021learning, cai2021appearance, cheng2023learning, cheng2024denoising} are exploited to analyze the extracted signals in the latent space from videos to achieve state-of-the-art performance. For instance, Huang \textit{et al.} \cite{huang2024long} effectively compensates for the information gap between the input sequence and the target frame by introducing the dynamic prototype alignment and long-distance frame prediction. Zhao \textit{et al.} \cite{zhao2025rethinking} rethink and try to improve those unsupervised VAD methods that are based on recurrent neural networks and their variants. However, there still remain unresolved issues that are suppressing the development of VAD.

First of all, there are actually different detection demands under various circumstances. However, current methods focus on increasing the performance of fixed requirements \cite{chang2020clustering, zhang2020normality, cheng2023spatial}. Most of their models need to be trained all over again to cope with the new demands. When the detection demand changes, it can be considered that the tolerance of different anomalies changes accordingly. Thus, we propose the Degree of Tolerance (DoT) to label different demands. For example, we may consider the use of all modes of transportation on the sidewalk as abnormal at first, which can be called DoT $1$. However, it can actually be subdivided into multiple categories. In some scenarios, only motor vehicles are strictly prohibited, while non-motor vehicles can pass through, which can be called DoT $2$. It will then lead to the inapplicability of the original model of DoT $1$. Thus, we propose to design the Configurable Video Anomaly Detection (CVAD) with flexible solutions as a systematical method to address this issue.

CVAD could provide suitable and flexible solutions in more scenarios. It is designed hierarchically. We can apply several sets of blocks to make configurable detections in different DoTs. They can assemble different abilities by expanding or contracting blocks to alter and gain the detection result in current settings. Moreover, CVAD only needs to train the newly added block with previous blocks frozen if a new DoT is proposed. Experiments also show that its training time is much less than training a new model from scratch to obtain the same detection result. Besides the configurable ability, our proposed CVAD can make full use of organizing multiple blocks to help improve detection accuracy. Blocks can gradually represent more and more complex patterns \cite{oreshkinn2020nbeats}, which means one block can extract information more deeply based on previous ones.

Another important point lies in that video sequences obviously have characteristics of consecutive changes with the potential to help VAD \cite{xu2024unified, nguyen2024hig}. Especially for abnormal objects, the interplay of context information is important. Spatial-temporal context information \cite{zhao2024learning, lin2024ffstie} needs more accurate selection and refinement. However, the appearance of objects can be anticipated to have enormous changes after a while \cite{xu2024boundary}. For instance, a movement from near to far means the size of the object is becoming smaller. If dynamically capturing the information of objects is limited, then the information passed in the chain form could be broken \cite{lou2025video}, resulting in false detections, making properly modeling coherence crucial. Inspired by \cite{li2024lsknet}, we propose the Multi-Scale Memory with Selective Mechanism (MS$^2$M) to boost CVAD's ability of dynamically exploring features. CVAD can better understand normality with the multi-scale normal patterns weighted by a series of memory features of abundant receptive fields. 

\begin{figure*}[t]
    \centerline{\includegraphics[width=\textwidth]{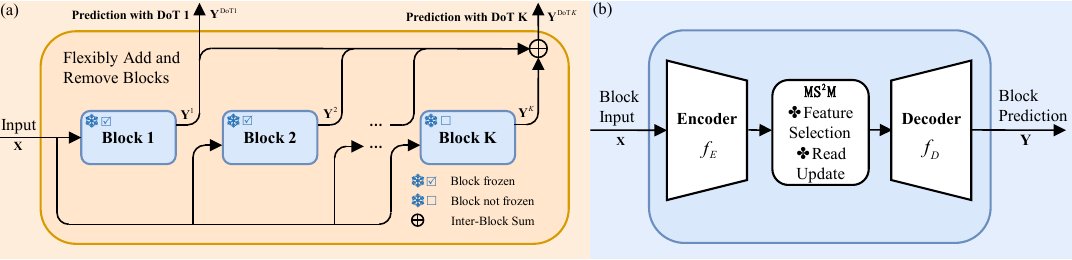}}
    \caption{The overall architecture of configurable video anomaly detection with flexible solutions. (a) and (b) represent the stack and block level, respectively.}
    \label{f1}
\end{figure*}

\begin{figure}[htb]
    \centerline{\includegraphics[width=.5\textwidth]{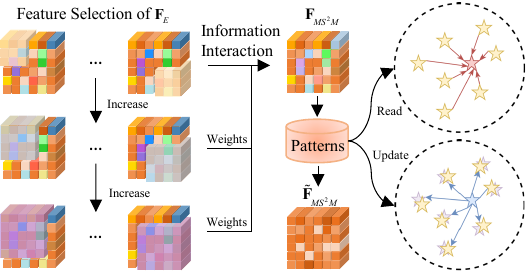}}
    \caption{The architecture of the multi-scale memory with selective mechanism.}
    \label{f2}
\end{figure}

\section{Methods}

\subsection{Configurable Video Anomaly Detection with Flexible Solutions}
As shown in Fig.~\ref{f1} (a), our proposed CVAD can be delineated into a two-tier hierarchy: the stack and block levels. We denote video frames as ${\bf X}_{n}, n\in\left[1,N\right]$. Then each data sample can be denoted as ${\bf X}_{n:n+t-1}$, where $t$ is the length of the sequence. At the stack level, it utilizes the double-nested connections to distribute data ${\bf X}_{n:n+t-2}$ into blocks and integrate the information of different blocks into a stack. As different DoT has distinct standards of discriminating normal and abnormal samples, the integration of the output of the first $k$ blocks is considered the prediction in DoT $k$:
\begin{equation}
    {\bf Y}^{\mathrm{DoT} \, k}_{n+t-1} = \sum^k_{k^{'}=1}{\bf Y}^{k^{'}}_{n+t-1},
\end{equation}
where ${\bf Y}^{\mathrm{DoT} \, k}_{n+t-1}$ is the prediction used for the detection in DoT $k$ and ${\bf Y}^{k^{'}}_{n+t-1}$ is the output of the $k^{'}$-th block. 

Thus, CVAD is configurable and flexible when facing the demands of different DoTs. For instance, when a new DoT $K+1$ is proposed while there have already been $K$ DoTs with their blocks well trained. We only need to freeze the previous blocks and add a new block after them with much less training time than training from scratch to obtain the detection result of DoT $K+1$. Moreover, such structures allow the following blocks to leverage the foundation of the previous ones and dig deeper, which will boost the detection performance even within one DoT. 

As shown in Fig.~\ref{f1} (b), the block is primarily comprised of three components: the encoder $f_E(\cdot)$, decoder $f_D(\cdot)$, and MS$^2$M. $f_E(\cdot)$ and $f_D(\cdot)$ are designed and inspired by \cite{ronneberger2015u} to seamlessly work with MS$^2$M. To simplify the expressions, we focus on introducing one block and omit irrelevant superscripts and subscripts. The encoder inputs ${\bf X}_{n:n+t-2}$ and outputs the feature ${\bf F}_E$. With the processing of our proposed MS$^2$M, $f_D(\cdot)$ finally inputs the fusion of ${\bf F}_E$ and ${\bf \tilde F}_{MS^2M}$ to generate the block prediction. Then the block predictions are integrated at the stack level as a crucial element to aid in VAD at the inter-block level. As our method adopts the prediction task as the proxy task to learn the normal patterns, it means we have the foundation to model the inherent coherence characteristics in videos. However, we need to ensure that the learned patterns can accurately represent the normality of coherence. Thus, we design MS$^2$M to fulfill the purpose.

\subsection{Multi-Scale Memory with Selective Mechanism}
As shown in Fig.~\ref{f2}, the output ${\bf F}_{E}$ of $f_E$ contains various information extracted from video frames. Although such information is already deeply embedded into latent space by our designed CVAD, the important coherence between consecutive frames still needs to be accurately captured. To achieve this, the MS$^2$M will establish the essential normal patterns from ${\bf F}_{E}$ to model the coherence. Overall, the function of MS$^2$M can be decomposed into two stages. It will first perform the Feature Selection (FS) of ${\bf F}_{E}$. As stated before, multi-scale memory patterns are beneficial to better perceiving the continuous changes of objects in terms of appearance, which is also highly connected to the FS stage. Specifically, we adopt a series of depth-wise convolutions to avoid directly using large kernel convolution. It can not only expand the receptive field rapidly but also prevent resource assumptions caused by large kernel convolution \cite{li2024lsknet}:
\begin{equation}
    \Omega_{j+1} = \Omega_{j}+\delta_{j+1}\cdot\left( \kappa_{j+1}-1 \right), j \in \left[1,J-1\right],
\end{equation}
where $\Omega$ is the receptive field, $\kappa_{j+1}$ and $\delta_{j+1}$ are the kernel size and dilation of $\left(j+1\right)$-th depth-wise convolution, and $J$ is the number of depth-wise convolutions. Then, we apply the weighted sum to results of each convolution, which will grant MS$^2$M the self-adaptive ability to dynamically select and refine key information according to the specific ${\bf F}_{E}$. To better learn and reinforce interactions of spatial-temporal context in coherence, we further employ $1\times1$ convolution after each layer. It will enhance the understanding of coherence in normal patterns:
\begin{equation}
    {\bf F}^{j+1}_{MS^2M} = f^{1\times1}_j\left(f_{\delta_j,\kappa_j}\left( {\bf F}^{j}_{MS^2M} \right)\right), {\bf F}^{1}_{MS^2M}={\bf F}_{E},
\end{equation}
where ${\bf F}^{j+1}_{MS^2M}$ is the output of the $j$-th operation, $f_{\delta_j,\kappa_j}\left(\cdot\right)$ represents the $j$-th depth-wise convolution, and $f^{1\times1}_j\left(\cdot\right)$ is the $1\times1$ convolution for its corresponding feature map. Therefore, we have the multi-scale features ${\bf F}^{set}_{MS^2M}=\left\{{\bf F}^{j }_{MS^2M}\right\}$. The selection and integration of these features are also important. To begin with, we apply the channel-based pooling to all of them in ${\bf F}^{set}_{MS^2M}$ and then generate their corresponding maps through a convolution denoted as $f_{attn}\left(\cdot\right)$:
\begin{equation}
    {\bf F}_{attn}=f_{attn}\left(Pool\left({\bf F}^{set}_{MS^2M}\right)\right).
\end{equation}
At last, the final result denoted as ${\bf F}_{MS^2M}$ of the FS stage is the fusion of selected features weighted by the calculated weights:
\begin{equation}
    {\bf F}_{MS^2M} = {\bf F}_E \odot f_{J}\left(\sum\limits^{J}_{j=1}{\bf F}^j_{attn}\cdot{\bf F}^{j}_{MS^2M}\right),
\end{equation}
where $f_{J}\left(\cdot\right)$ is a convolution served as the last fusion of this multi-scale information and $\odot$ represents Hadamard product.

\begin{table}[t]
    \centering
    \caption{Results of the quantitative frame-level AUC comparison. Bold numbers indicate the best performances and underscored ones are the second best.
    }
    \label{t1}
    \resizebox{.48\textwidth}{!}{
    \begin{tabular}{@{}lccc@{}}
    \toprule
    \textbf{Methods} & \textbf{UCSD Ped2} & \textbf{CUHK Avenue} & \textbf{ShanghaiTech} \\ \midrule
    FFP \cite{liu2018future}   & 95.4\% & 84.9\% & 72.8\% \\
    MemAE \cite{gong2019memorizing}  & 94.1\% & 83.3\% & 71.2\% \\
    MNAD \cite{park2020learning} & 97.0\% & 88.5\% & 70.5\% \\
    STCEN \cite{hao2022spatiotemporal} & 96.9\% & 86.6\% & 73.8\%  \\
    AMAE \cite{liu2022appearance} & 97.4\% & 88.2\% & 73.6\%  \\
    STD \cite{chang2022video} & 96.7\% & 87.1\% & 73.7\%  \\
    Zhong \textit{et al.} \cite{zhong2022cascade} & 97.7\% & 88.9\% & 70.7\%  \\
    Le \textit{et al.} \cite{le2023attention}  & 97.4\% & 86.7\% & 73.6\%  \\
    USTN-DSC \cite{yang2023video}  & \underline{98.1\%}  & \underline{89.9\%}  & 73.8\%  \\
    PDM-Net \cite{huang2024long}  & 97.7\%  & 88.1\%  & \underline{74.2\%}  \\
    LGN-Net \cite{zhao2025rethinking}  & 97.1\%  &89.3\%  & 73.0\% \\ \midrule
    CVAD (Ours)  & \textbf{99.2\%}  & \textbf{90.7\%}  & \textbf{74.9\%}  \\ \bottomrule
    \end{tabular}}
\end{table}

In the second stage, we first organize the ${\bf F}_{MS^2M}$ as a set of queries denoted as $\left\{{\bf f}_{MS^2M} \in \mathbb{R}^{M \times C}\right\}$, where $M$ is the number of queries and $C$ is the number of channels. We denote ${\bf f}^m_{MS^2M} \in \mathbb{R}^{C}$ as individual queries in the query set. Firstly, we utilize the query set to read the normal patterns and reconstruct ${\bf \tilde f}^m_{MS^2M}$. Then we update the normal patterns to memorize latent representations of normality. The normal patterns consist of $N$ latent vectors that implicitly represent the normality distribution. We denote the vectors as $k^n \in \mathbb{R}^C$. In the reading phase, we first compute the attention scores between ${\bf f}^m_{MS^2M}$ and all normal patterns as follows:
\begin{equation}
    a^{m,n} = \frac{\exp\left( {\left( {{\bf f}^m_{MS^2M}} \right)}^T k^n \right)}
    {\sum^N_{n^{'}=1}\exp\left( {\left( {{\bf f}^m_{MS^2M}} \right)}^T k^{n^{'}} \right)}.
\end{equation}
For each query ${\bf f}^m_{MS^2M}$, we apply a weighted average of all patterns with corresponding attention scores to reconstruct as follows:
\begin{equation}
    {\bf \tilde f}^m_{MS^2M} = \sum\limits^N_n a^{m,n} k^n.
\end{equation}
Thus, we apply the same operations to all queries and organize them to obtain the final feature ${\bf \tilde F}_{MS^2M}$. As for the update of the normal patterns, we compute the attention scores between $k^n$ and all queries similarly. Then we apply a weighted average of all queries with corresponding attention scores as the increment and apply a sigmoid function to update patterns. Overall, the loss function of CVAD can be formulated as:
\begin{equation}
    \resizebox{.43\textwidth}{!}{$
    \begin{aligned}
        \mathcal{L} &= \sum\limits^M_{m=1} \left( \left\Vert {\bf f}^m_{MS^2M}-k^{m\rightarrow1} \right\Vert^2_2 - \left\Vert {\bf f}^m_{MS^2M}-k^{m\rightarrow2} \right\Vert^2_2 \right) \\
                    &+ \left\Vert {\bf Y}-{\bf X} \right\Vert^2_2
    \end{aligned},
    $}
\end{equation}
where $k^{m\rightarrow1}$ and $k^{m\rightarrow2}$ are the first and second nearest patterns to ${\bf f}^m_{MS^2M}$, respectively.

\section{Experiments}

\begin{figure}[t]
    \centerline{\includegraphics[width=.5\textwidth]{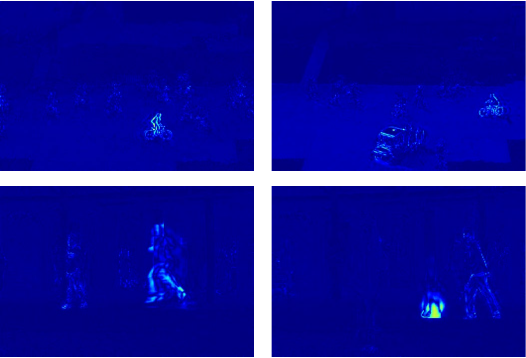}}
    \caption{Illustration of the difference of the predicted frames and ground truth.}
    \label{f3}
\end{figure}

\subsection{Detection Criterion and Implementation Details}
During the inference phase, we detect the samples based on abnormal scores calculated as follows:
\begin{equation}
    \resizebox{.43\textwidth}{!}{
    $e = g\left( \tau \cdot PSNR\left({\bf Y}, {\bf X} \right) + \left(1-\tau\right) \cdot \left\Vert {\bf F}_{MS^2M}-{\bf \tilde F}_{MS^2M} \right\Vert^2_2 \right)$
    },
\end{equation}
where $e$ is the abnormal score and $g\left(\cdot\right)$ is the normalization and filtering function.

We conduct sufficient experiments on three commonly used benchmarks: UCSD Ped2 \cite{ped2}, CUHK Avenue \cite{avenue}, and ShanghaiTech datasets \cite{shanghaitech}. As stated before, we further discuss the applicability of motor and non-motor vehicles in the scenario to show the distinct advantages of our CVAD by proposing a new dataset with great compatibility called ped2DoT based on the ped2 dataset. It provides a feasible and convenient platform for future research interested in the flexibility of VAD. For now, ped2DoT contains two basic DoTs. Assuming both motor and non-motor vehicles are prohibited, we call this situation DoT $1$ with the strictest settings. We determine the situation that non-motor vehicles are allowed to be DoT $2$, which brings great challenges to the DoT $1$ model. Moreover, the frame-level Area Under the Curve (AUC) \cite{cheng2024normality} is chosen as the evaluation metric. We use an NVIDIA GeForce RTX 4090 GPU with a batch size of 6 for model training, which is implemented by PyTorch \cite{paszke2019pytorch}.

\begin{figure}[t]
    \centerline{\includegraphics[width=.5\textwidth]{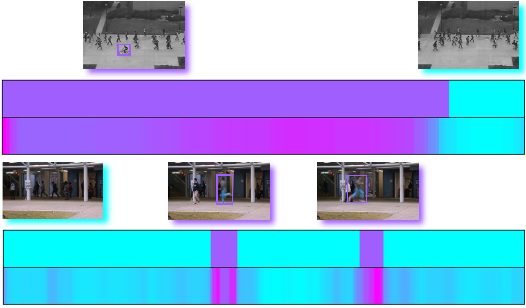}}
    \caption{Illustration of temporal localization results of the detection.}
    \label{f4}
\end{figure}

\begin{figure}[t]
    \centerline{\includegraphics[width=.5\textwidth]{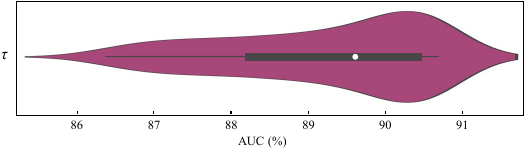}}
    \caption{The qualitative analysis of hyperparameter $\tau$.}
    \label{f5}
\end{figure}

\subsection{Comparison with Previous Methods}
As shown in Table~\ref{t1}, CVAD exhibits its outstanding detecting ability by comparing with previous methods, exceeding the second best performance of 1.1\%, 0.8\%, and 0.7\% on three benchmark datasets. Compared with methods that try to model normal patterns \cite{gong2019memorizing, park2020learning, chang2022video, yang2023video}, we obtain improvements of 1.1\% - 5.1\%, 0.8\% - 7.4\%, and 1.1\% - 4.4\% on three benchmarks, fully proving the effectiveness of our proposed MS$^2$M. As for the ability of spatial-temporal modeling, the result shows that our CVAD is 1.5\%, 1.8\%, and 1.1\% higher than those methods \cite{hao2022spatiotemporal, liu2022appearance, zhong2022cascade} specifically designed for spatial-temporal explorations. The competitive quantitative results are obtained based on the notable coherence modeling ability of our CVAD, showing that our proposed multi-scale spatial-temporal features indeed help better capture anomalies, especially those caused by abnormal incoherence. 

\subsection{Visualization Results}
The visual results of the difference of predicted frames and their corresponding ground truth are shown in Fig.~\ref{f3}. The anomalies are clear and obvious by observing these error maps while normal parts are maintained and reconstructed with good quality. It further illustrates CVAD's excellent understanding of videos and sets a better detection boundary for VAD. Moreover, the results also show that various abnormal objects that have different appearance characteristics (\textit{e.g.} size and orientation) are accurately captured thanks to our CVAD's automatic adjusting ability.

Fig.~\ref{f4} shows the temporal analysis of abnormal scores of consecutive videos. It can be observed that abnormal scores are steady and accurate with no huge fluctuations. The conversion between normal and abnormal detections is also smooth and acute. The results fully show CVAD's understanding of video coherence. Thanks to MS$^2$M, CVAD can select and refine important information of embeddings in the latent space with better coherence modeling, which involves removing noises, integrating key features, and enhancing the understanding of the context. What's more, we also conduct the sensitivity analysis on the hyperparameter $\tau$ on the CUHK Avenue dataset as shown in Fig.~\ref{f5}. It can be observed that most model results can still obtain rather high performances within the acceptable range, proving the robustness of our proposed CVAD.

\subsection{Analysis of CVAD's flexible solutions}
Table~\ref{t2} shows the distinct advantages of CVAD in terms of its configurability and flexibility. We conduct several experiments by using strategies of both CVAD and training from scratch. By freezing previous blocks, it only takes 2 epochs for CVAD to train the newly added block and obtain the highest AUC with 99.3\%, which is Model \Rmnum{1}. However, training from scratch has to take 20$\times$ time than CAVD to obtain an AUC with 99.1\%, which is Model \Rmnum{5}. Not to mention performances of Model \Rmnum{2}-\Rmnum{4} trained with 2, 10, and 20 epochs take 1$\times$, 5$\times$, 10$\times$ time than CVAD are far from satisfying. Gaps of performances between the two strategies are also listed in Table~\ref{t2}, further confirming the resource-saving ability of CVAD.

\begin{table}[t]
    \centering
    \caption{The AUC and training time comparison of CVAD and training new models from scratch on the ped2DoT2 dataset. Bold numbers indicate the best performances with the least time.}
    \label{t2}
    \newcolumntype{C}{>{\centering\arraybackslash}X}      
    \newcolumntype{L}[1]{>{\centering\arraybackslash}p{#1}}
    \small
    \begin{tabularx}{.48\textwidth}{L{1.5cm}CCCCC}
    \toprule
    \multicolumn{2}{c}{\textbf{Model}}  & \textbf{Epochs}  & \textbf{(Times)}  & \textbf{AUC}  & \textbf{Gap} \\ \hline
    \rowcolor[gray]{.9}
    CVAD  & \Rmnum{1}  & \textbf{2}  & \textbf{(1$\times$)}  & \textbf{99.3\%}  & N/A \\
    \multirow{4}{*}{\makecell[c]{Train from \\ scratch}}  & \Rmnum{2}  & 2  & (1$\times$)  & 84.7\%  & 14.6\%  \\
    & \Rmnum{3}  & 10  & (5$\times$)  & 82.8\%  & 16.5\%  \\
    & \Rmnum{4}  & 20  & (10$\times$)  & 93.4\%  & 5.9\%  \\
    & \Rmnum{5}  & 40  & (20$\times$)  & 99.1\%  & 0.2\%  \\ \bottomrule
    \end{tabularx}
\end{table}

\section{Conclusion}
In conclusion, we propose CVAD to cope with VAD demands that actually vary from time to time. It is configurable and able to output flexible solutions when facing different DoTs. We design a hierarchical structure, where blocks continuously provide valuable insights for the following ones to better retrieve normal patterns of spatial-temporal coherence. Moreover, the proposed MS$^2$M selects and refines key information with a dynamic adjusting ability to model multi-scale features of coherence. Experiments on three benchmark datasets and extended analysis demonstrate that CVAD has an outstanding detecting and reliable configurable capability.

\bibliographystyle{IEEEbib}


\end{document}